\documentclass[runningheads]{llncs}

\usepackage{eccv}

\usepackage{eccvabbrv}

\usepackage{graphicx}
\usepackage{booktabs}

\usepackage[accsupp]{axessibility}  % Improves PDF readability for those with disabilities.

\definecolor{cvprblue}{rgb}{0.21,0.49,0.74}
\definecolor{darkgreen}{rgb}{0.0, 0.5, 0.0}
\definecolor{darkred}{rgb}{0.55, 0.0, 0.0}
\definecolor{deeppeach}{rgb}{1.0, 0.8, 0.64}
\definecolor{lightgray}{rgb}{0.83, 0.83, 0.83}

\newcommand{\DatasetName}{MVSign\xspace}

\newcommand{\TitleName}{PHOSA\xspace}

\newcommand{\SamplingName}{Motion-aware Data Sampling Strategy\xspace}

\usepackage{url}            % simple URL typesetting
\usepackage{booktabs}       % professional-quality tables
\usepackage{amsfonts}       % blackboard math symbols
\usepackage{nicefrac}       % compact symbols for 1/2, etc.
\usepackage{microtype}      % microtypography
\usepackage{xcolor}         % colors

\usepackage{float}
\usepackage{wrapfig}
\usepackage{caption}
\usepackage{multirow}
\usepackage{multicol}
\usepackage{bbding}
\usepackage{makecell}
\usepackage{amsmath}
\usepackage{enumitem}

\DeclareMathOperator*{\argmin}{arg\,min}
\usepackage[ruled,vlined]{algorithm2e}
\usepackage{algpseudocode}
\usepackage{graphicx}
\usepackage{threeparttable}

\usepackage{booktabs}
\usepackage{subcaption}

\usepackage[pagebackref,breaklinks,colorlinks,citecolor=eccvblue]{hyperref}

\usepackage{orcidlink}

\begin{document}

\title{
\TitleName: Photorealistic 3D Sign Avatar Modeling and Benchmark}

\author{Haodong Wang\inst{1} \and
Hezhen Hu\inst{2*} \and
Wengang Zhou\inst{1*} \and
Houqiang Li\inst{1}}

\authorrunning{H.~Wang et al.}

\institute{University of Science and Technology of China \and
University of Texas at Austin}

\maketitle
\begin{center}
   \centering
   \captionsetup{type=figure}
   \includegraphics[width=\linewidth]{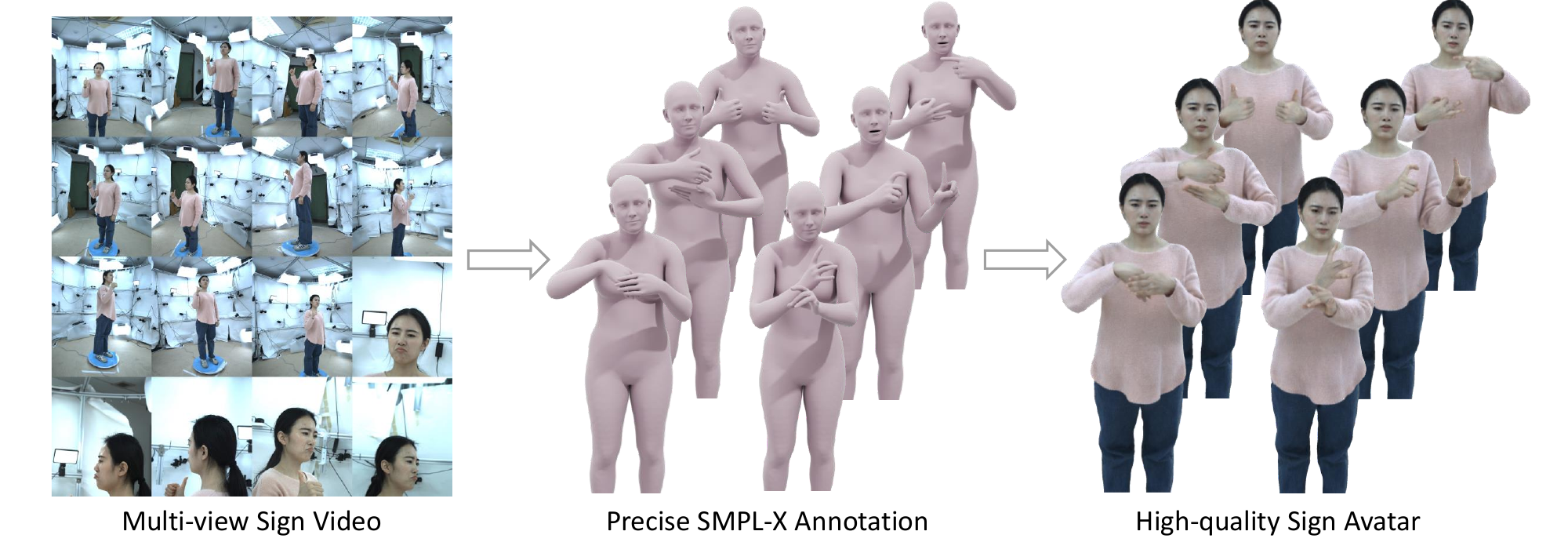}
    \captionof{figure}{Overview of \textbf{\TitleName} for photorealistic 3D sign avatar modeling. To deal with sign avatar modeling, we present: (1)\textbf{\DatasetName} (left), a multi-view Chinese sign language benchmark captured by 16 synchronized RGB cameras (2048$\times$2448), together with precise hand and facial annotations (middle) produced by our hybrid SMPL-X fitting pipeline; (2): Photorealistic sign avatar (right) generated from \DatasetName via our decoupled sign avatar representation, showing high fidelity and generalization to complex sign articulations.}
   \label{fig:teaser}
\end{center}%

\newcommand{\nonumberfootnote}[1]{%
  \renewcommand{\thefootnote}{}%
  \footnotetext{\hspace*{-0.9em}#1}
  \renewcommand{\thefootnote}{\arabic{footnote}}
}

\nonumberfootnote{* Corresponding authors.}

\begin{abstract}
In this work, we focus on photorealistic sign avatar modeling, which is crucial for effective communication with the Deaf community and is characterized by complex hand gestures and nuanced facial expressions.
To this end, we introduce \DatasetName, the first multi-view Chinese sign language dataset co-designed with Deaf experts, featuring diverse gestures and rich annotations. 
For precise SMPL-X annotation, we develop a hybrid fitting pipeline that produces accurate body, hand, and facial parameters and can also be applied to the monocular setting.
Building on \DatasetName, we propose a decoupled sign avatar representation that isolates body, head, and hand components to capture complex articulations, together with a motion-aware sampling strategy to handle motion blur and balance gesture diversity.
Extensive experiments demonstrate that our method achieves high-fidelity visual results on \DatasetName, particularly in detailed hand and facial regions, and generalizes well to in-the-wild monocular sign language videos. Project page: \url{https://naaapi.github.io/PHOSA}.
\end{abstract}
    
\section{Introduction}
Sign Language Production (SLP), which converts spoken language into continuous sign sequences, is crucial for facilitating communication between hearing and Deaf community.
While recent SLP works~\cite{yu2024signavatars, zuo2024simple, baltatzis2024neural} have achieved progress in motion generation, they typically represent signers using parametric body meshes such as SMPL-X~\cite{SMPL-X:2019}. These mesh-based representations lack photorealism and fine-grained expressiveness, failing to deliver the natural, human-like signing experiences that Deaf community strongly prefer~\cite{quandt2022attitudes}.
This highlights the need to develop photorealistic and animatable avatars with SMPL-X parameter-driven articulation capabilities.

In this work, we focus on photorealistic and drivable sign avatar modeling.
Our goal is to build the dataset, annotation pipeline, and representation needed to faithfully capture expressive signing.
Achieving this goal is challenging for two main reasons. First, suitable datasets are lacking. Existing multi-view human datasets~\cite{cheng2023dna, icsik2023humanrf, peng2021neural, xiong2024mvhumannet} mainly capture general body motion and overlook the detailed hand articulations and facial expressions essential for sign communication. Conversely, current sign language datasets~\cite{yu2024signavatars, PHOENIX14t, Duarte_CVPR2021} are designed for motion production, without the multi-view imagery or precise annotations needed for human avatar modeling. Second, modeling fine-grained articulations is difficult. 
Prior avatar modeling works~\cite{lei2024gart, li2024animatablegaussians, shao2024splattingavatar, hu2024gaussianavatar} target general human motions and struggle with modeling on complex hand gestures and facial expressions.

To overcome these challenges, we introduce \DatasetName (see Figure~\ref{fig:teaser}), the first multi-view Chinese sign language dataset co-designed with Deaf experts and collected under IRB approval. 
It covers diverse gestures spanning basic hand shapes and complex motion patterns, supporting generalization to unseen signs. 
Each sample includes rich annotations, \emph{e.g.,} image matting, body-part segmentation, 3D keypoints, and SMPL-X parameters. 
To ensure accurate recovery of expressive details, we develop a hybrid SMPL-X fitting pipeline that integrates predictions from multiple state-of-the-art models, achieving precise hand articulation and nuanced facial motion. The pipeline can also serve as a plug-in for monocular RGB settings.

To deal with fine-grained articulation during sign avatar modeling, we further develop an efficient decoupled sign avatar representation that separates the whole-body Gaussian maps into body, head, and hands. 
By introducing partial kinematic decoupling, body joints are fixed while hand mobility is preserved, making hand pose maps independent of body motion and improving gesture generalization. 
Furthermore, a motion-aware frame sampling strategy filters motion-blurred frames and balances diverse motion types, enhancing visual fidelity and training stability.
Extensive experiments demonstrate that our method achieves state-of-the-art performance on \DatasetName and in-the-wild sign videos from the Web.

In summary, our main contributions are threefold:
\begin{itemize}[leftmargin=*, itemsep=0pt]
    \item We systematically study photorealistic sign avatar modeling and establish a multi-view benchmark to facilitate evaluation and comparison.
    \item We present \DatasetName, the first multi-view Chinese sign language dataset co-designed with Deaf experts, featuring diverse gestures and rich annotations. To ensure annotation accuracy, we develop a hybrid SMPL-X fitting pipeline to achieve precise hand and facial parameters.
    \item We propose a decoupled avatar representation that separates body, head, and hands, and employs partial kinematic decoupling for improved gesture generalization and fine-grained articulation fidelity.
\end{itemize}
\section{Related Work}
\label{sec:relatedwork}

\subsection{Animatable Human Avatar Modeling}
\label{sec:rel_avatar}

Model animatable human avatar from RGB video is a long-standing and challenging problem. Early work~\cite{weng_humannerf_2022_cvpr, icsik2023humanrf, Zhao_2022_CVPR, jiang2022neuman, wang2023styleavatar, liu2021neural, guo2023handnerf, hong2022headnerf, jiang2023instantavatar} uses implicit neural representations such as NeRF~\cite{mildenhall2021nerf} or SDF to model human avatar. However, they struggle to balance high-quality results with fast rendering. Recently, 3D Gaussian Splatting~\cite{kerbl20233d} has made substantial progress in improving both training and rendering times over traditional NeRFs while preserving high quality, inspiring many works~\cite{lei2024gart, hu2024gaussianavatar, li2024animatablegaussians, hu2024expressive, moon2024exavatar, Pang_2024_CVPR, zielonka25dega} to adopt it as a foundational representation.
GART~\cite{lei2024gart} uses a mixture of moving 3D Gaussians to explicitly approximate the geometry and appearance of deformable subjects, leveraging categorical template models with learnable forward skinning. 
EVA~\cite{hu2024expressive} introduces a plug-and-play module that significantly ameliorates SMPL-X misalignment issues, while a context-aware adaptive density control strategy is applied to accommodate the varied granularity across body parts.
AnimatableGaussians~\cite{li2024animatablegaussians} learns a parametric template from the input videos, and then parameterizes the template on canonical Gaussian maps where each pixel represents a 3D Gaussian. Such template-guided 2D parameterization enables them to employ a powerful StyleGAN~\cite{karras2019style}-based CNN to learn the pose-dependent Gaussian maps for modeling detailed dynamic appearances.

Different from previous works, which mainly focus on general human motions, our network introduces an efficient decoupled sign avatar representation specifically designed to capture fine-grained hand and facial details essential for photorealistic sign avatar modeling.

\subsection{Human-centric and Sign Language Benchmarks}
\label{sec}

Human-centric benchmarks provide the foundation for learning and evaluating photorealistic, animatable human representations.
Recent multi-view human datasets, such as DNA-Rendering~\cite{cheng2023dna}, HumanRF~\cite{icsik2023humanrf}, ZJU-MoCap~\cite{peng2021neural}, and MVHumanNet~\cite{xiong2024mvhumannet}, have advanced high-fidelity novel-view synthesis and animatable avatar modeling.
However, they mainly focus on general human motions and do not specifically target the complex hand articulations and subtle facial expressions required for sign communication.

In parallel, existing sign language datasets mainly focus on recognition, translation, or motion production.
For example, PHOENIX14T~\cite{PHOENIX14t}, CSL-Daily~\cite{csldaily} and How2Sign~\cite{Duarte_CVPR2021} provide valuable video-language annotations for sign understanding, while SMILE~\cite{ebling2018smile} targets lexical-level sign recognition and assessment. SignAvatars~\cite{yu2024signavatars} further introduces large-scale 3D motion annotations for sign language production.
Nevertheless, these datasets are not designed for photorealistic sign avatar reconstruction. Most lack synchronized multi-view imagery and annotations required by human avatar modeling benchmarks.

This gap motivates our \DatasetName benchmark, which provides multi-view sign videos together with rich annotations and a dedicated annotation pipeline, enabling systematic evaluation of photorealistic sign avatar modeling.
\section{\DatasetName Dataset}
\label{sec:method}

\begin{table}[t]
\centering
\tiny
\caption{Dataset statistics comparison with existing multi-view human-centric datasets and sign language datasets.}
\begin{threeparttable}
    \resizebox{\linewidth}{!}{%
        \begin{tabular}{l|ccc|ccc|c}
    \toprule
    Dataset & Sign & Head & SMPL-X & \#View & \#ID & \#Frames & Resolution\\
    \midrule
    PHOENIX14T~\cite{PHOENIX14t} & \textcolor{darkgreen}{\CheckmarkBold} & \textcolor{darkred}{\XSolidBrush} & \textcolor{darkred}{\XSolidBrush} & 1 & 9 & 0.94M & 260P \\
    CSL-Daily~\cite{csldaily} & \textcolor{darkgreen}{\CheckmarkBold} & \textcolor{darkred}{\XSolidBrush} & \textcolor{darkred}{\XSolidBrush} & 1 & 10 & 1.5M & 512P \\
    How2Sign~\cite{Duarte_CVPR2021} & \textcolor{darkgreen}{\CheckmarkBold} & \textcolor{darkred}{\XSolidBrush} & \textcolor{darkred}{\XSolidBrush} & 2 & 11 & 5.7M & 720P \\
    SignAvatars~\cite{yu2024signavatars} & \textcolor{darkgreen}{\CheckmarkBold} & \textcolor{darkred}{\XSolidBrush} & \textcolor{darkgreen}{\CheckmarkBold} & 1 & 153 & 8.34M & - \\
    SMILE~\cite{ebling2018smile} & \textcolor{darkgreen}{\CheckmarkBold} & \textcolor{darkred}{\XSolidBrush} & \textcolor{darkred}{\XSolidBrush} & 4 & 66 & - & 1080P \\
    VSL\tnote{*}~\cite{vsl} & \textcolor{darkgreen}{\CheckmarkBold} & \textcolor{darkred}{\XSolidBrush} & \textcolor{darkred}{\XSolidBrush} & 16 & 2 & 50K & 4096P \\
    SGNify\tnote{*}~\cite{forte2023reconstructing} & \textcolor{darkgreen}{\CheckmarkBold} & \textcolor{darkred}{\XSolidBrush} & \textcolor{darkgreen}{\CheckmarkBold} & 12 & 3 & - & - \\
    Human3.6M~\cite{h36m_pami} & \textcolor{darkred}{\XSolidBrush} & \textcolor{darkred}{\XSolidBrush} & \textcolor{darkgreen}{\CheckmarkBold} & 4 & 11 & 3.6M & 1000P \\
    NHR~\cite{wu2020multi} & \textcolor{darkred}{\XSolidBrush} & \textcolor{darkred}{\XSolidBrush} & \textcolor{darkgreen}{\CheckmarkBold} & 80 & 3 & 100K & 2048P \\
    ZJU-MoCap~\cite{peng2021neural} & \textcolor{darkred}{\XSolidBrush} & \textcolor{darkred}{\XSolidBrush} & \textcolor{darkgreen}{\CheckmarkBold} & 24 & 10 & 180K & 1024P \\
    THuman 4.0~\cite{zheng2022structured} & \textcolor{darkred}{\XSolidBrush} & \textcolor{darkred}{\XSolidBrush} & \textcolor{darkgreen}{\CheckmarkBold} & 24 & 3 & 10K & 1150P \\
    \midrule
    \DatasetName (Ours) & \textcolor{darkgreen}{\CheckmarkBold} & \textcolor{darkgreen}{\CheckmarkBold} & \textcolor{darkgreen}{\CheckmarkBold} & 16 & 5 & 115K & 2048P \\
    \bottomrule
\end{tabular}
    }
\begin{tablenotes}
\small
\item[*] The dataset is not publicly available.
\end{tablenotes}
\end{threeparttable}
\vspace{-2mm}
\label{tab:statistics}
\end{table}

\subsection{Data Capture}
\label{sec:dataset}

\noindent \textbf{Data Collection Protocol.}
As described by the Hamburg Notation System (HamNoSys)~\cite{hanke2004hamnosys}, sign language consists of several fundamental components, including hand shapes, movements, orientations, locations, and facial expressions. 
To ensure that \DatasetName covers diverse sign articulations, we collaborated with Deaf experts to carefully design the signing actions. 
First, we adopt 109 basic hand shapes, consisting of 26 letters, 22 counting numbers and 61 commonly used words from the reference book Chinese Sign Language Tutorial~\cite{ni2020zhong}, which is designed for daily sign-language learning and communication. 
These words represent the most frequently used signs in daily communication, forming a strong foundation for basic hand shapes. 
Next, to provide diverse hand motion patterns, facial expressions, and natural transitions between consecutive signs, we curated daily-use sign language sentences from the beginner and intermediate volumes of the Chinese Sign Language Tutorial. The books cover 20 common daily-life scenarios, including greetings, study, family, hobbies, weather, shopping, travel, and medical services.
Each scenario contains approximately 10 sentences.
For each recorded signer, we randomly selected 5 scenarios from the 20 scenarios and used the corresponding sentences, resulting in around 50 distinct sentences per signer.
This collection protocol ensures that \DatasetName captures both fundamental hand shapes and complex dynamic gestures, thereby enhancing the generalization ability of avatars to unseen sign poses.

\noindent \textbf{System Setup.} 
Our capture system consists of a high-fidelity array of 16 synchronized RGB cameras, each recording at a resolution of 2048×2448 pixels and 25 frames per second. 
The camera layout is carefully designed to optimize the modeling of sign avatars by balancing body coverage and fine-grained detail. 
Specifically, one camera captures the frontal view, ten cameras are distributed along the lateral arc to capture multi-angle side perspectives and detailed hand and arm articulations, while the remaining five focus on the head region to preserve fine facial expressions and mouthing. 
All cameras are temporally synchronized to ensure consistent multi-view capture, and the entire system is calibrated using VGGSfM~\cite{wang2024vggsfm}.

\noindent \textbf{Dataset Statistics and Comparison.} 
\DatasetName includes five native Chinese Sign Language signers (two males, three females) with high fluency and expressivity. 
Each actor contributed approximately 15 minutes of recording, totaling around 23,000 frames, with 10 minutes on basic hand shapes and 5 minutes on diverse sentences.
The detailed comparison between \DatasetName and other relevant datasets is shown in Table~\ref{tab:statistics}. 
Most existing sign language datasets~\cite{PHOENIX14t, csldaily, Duarte_CVPR2021, yu2024signavatars} used for production are monocular, short in duration, and relatively low in visual quality, which limits their suitability for benchmarking multi-view avatar modeling. In contrast, current multi-view human datasets~\cite{cheng2023dna, icsik2023humanrf, peng2021neural, xiong2024mvhumannet} primarily focus on general body motion and do not emphasize the fine-grained hand articulations and facial expressions that are critical for sign language.
\DatasetName is the only dataset that simultaneously provides multi-view sign language data, dedicated head portrait views, and SMPL-X parameters annotations, while also maintaining high resolution and a comparable number of frames.

\noindent \textbf{Ethics Considerations.} 
This project received IRB approval, and all participants provided informed written consent for public release of anonymized data. The IRB documentation will be released alongside the dataset.

\subsection{Data Annotation}
\label{sec:annotation}
To facilitate sign avatar modeling, our dataset contains comprehensive annotations along with the raw data, including image matting, body part segmentation, 3D keypoints, and SMPL-X parameters.

\noindent \textbf{Matting and Body Part Segmentation.}
We utilize the state-of-the-art human foundation model Sapiens~\cite{khirodkar2025sapiens} to extract body-part segmentation on the raw images. 
However, we observe that the segmented hand and hair regions often contain extraneous background areas. To address this issue, we further apply the SAM model~\cite{kirillov2023segany} to obtain refined human matting and remove unwanted background artifacts, ensuring clean and accurate body-part segmentation.

\begin{figure*}[t] \centering
    \includegraphics[width=\textwidth]{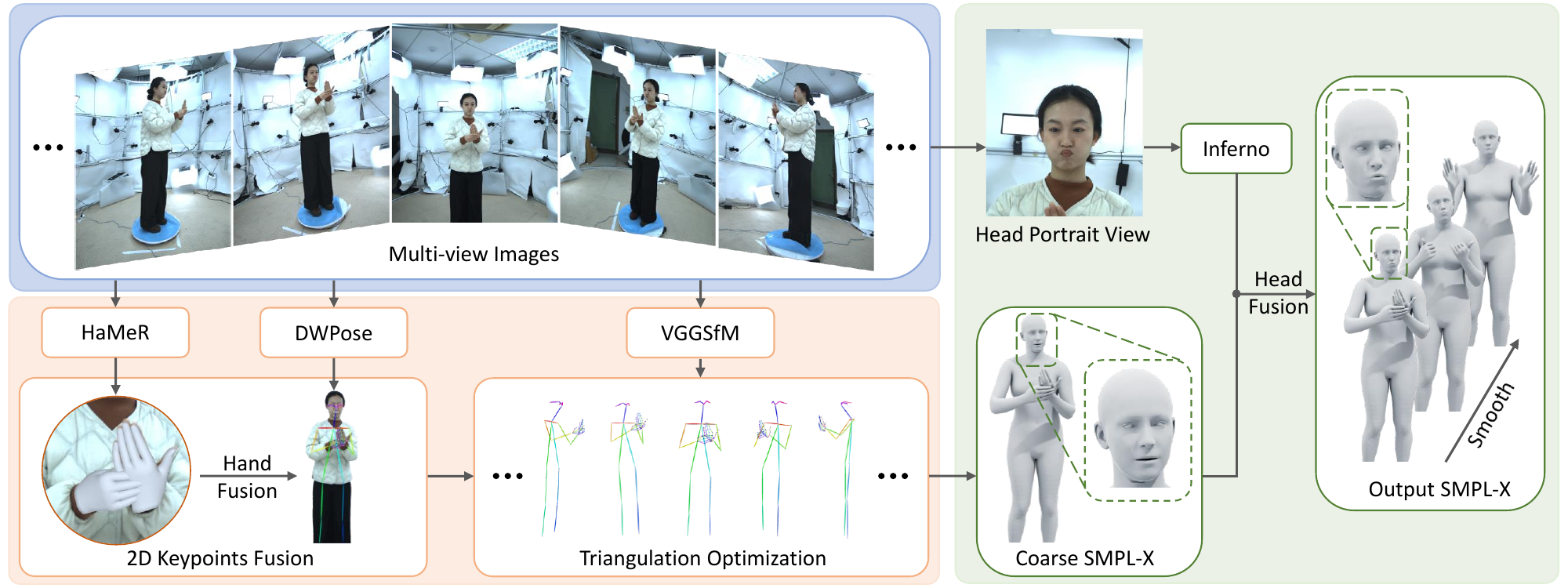}
    \caption{Overview of the hybrid SMPL-X fitting pipeline. Our pipeline fuses outputs from multiple models to leverage their complementary strengths, achieving precise and robust SMPL-X parameter estimation.} \label{fig: annotation_pipeline}
\vspace{-4mm}
\end{figure*}

\noindent \textbf{Keypoints and SMPL-X Parameters.}
Current whole-body pose estimation~\cite{cao2017realtime, yang2023effective, mmpose2020} and mesh recovery~\cite{lin2023osx, cai2023smplerx, Moon_2022_CVPRW_Hand4Whole} methods struggle to accurately capture complex sign language gestures. This motivates us to propose a \emph{hybrid SMPL-X fitting procedure} that integrates outputs from multiple state-of-the-art models. By leveraging the strengths of each model, our approach achieves precise 3D keypoint localization and SMPL-X parameter estimation. The full pipeline is shown in Figure~\ref{fig: annotation_pipeline}.

Given synchronized multi-view images $\mathcal{I} = \{I_1, ..., I_N\}$  with $N$ calibrated cameras. We first apply DWPose~\cite{yang2023effective} for each view to detect 2D keypoints $J^{2D}$, including keypoints of the body, hands and face. 
To improve hand mesh precision, we subsequently process each view through HaMeR~\cite{pavlakos2024reconstructing}, obtaining more robust MANO parameters~\cite{MANO:SIGGRAPHASIA:2017} and corresponding hand keypoints $J^{2D}_{h}$, which replace the original DWPose hand estimations and produce optimized whole-body keypoints $\hat{J}^{2D}$.
Leveraging calibrated camera intrinsics $\mathbf{K}$ and extrinsics $[\mathbf{R}|\mathbf{t}]$, we obtain precise 3D keypoints $\hat{J}^{3D}$ by applying multi-view geometric constraints via triangulation optimization~\cite{easymocap}:
\begin{equation}
\label{2dlift3d}
\hat{J}^{3D} = \argmin_{J^{3D}} \sum_{i=1}^N \left\|(\mathbf{K}_i[\mathbf{R}_i|\mathbf{t}_i]J^{3D}) - \hat{J}^{2D}_i\right\|_2.
\end{equation}
Then we fit SMPL-X parameters by minimizing the following objective:
\begin{equation}
\begin{aligned}
\mathcal{L}(\theta,\beta,\psi) = &\sum_{i \in \mathcal{J}} \gamma_i\omega_i \rho\left(R_{\theta}(J(\beta)_i) - \hat{J}^{3D}_{i}\right) +  \lambda_{\beta}\left\|\beta\right\|_2 + \lambda_{\psi}\left\|\psi\right\|_2,
\end{aligned}
\label{eq:fit_loss_term }
\end{equation}
where $\theta\in \mathbb{R}^{165}$ denotes the full body pose parameters, $\beta \in \mathbb{R}^{10}$ refers to body shape parameters, $\psi \in \mathbb{R}^{10} $ is the facial expression parameters, $\rho(\cdot)$ denotes the Geman-McClure robust kernel which helps prevent the disturbance from noisy supervision signals, $R_\theta(\cdot)$ denotes the function which rotates the $J(\beta)$ given the pose $\theta$, $\gamma$ and $\omega$ are predefined joint weights and confidence weights from detection.

Since the face keypoints are too sparse to capture nuanced facial articulations, the fitted facial parameters often fail to accurately represent the true expression. To address this, we utilize INFERNO~\cite{zielonka2022mica, EMOCA:CVPR:2021, filntisis2022visual, Feng:SIGGRAPH:2021} to extract more precise and expressive facial expression parameters and replace the initial facial parameters derived from skeletal fitting.

We further enforce temporal consistency of the SMPL-X parameters through two strategies: (1) employing temporal continuity constraints by initializing each frame's optimization with the preceding frame's converged parameters, and (2) applying SmoothNet~\cite{zeng2022smoothnet} to the resulting SMPL-X sequence to mitigate high-frequency artifacts.

\section{Sign Avatar Modeling}
\label{sec:avatar}
\begin{figure*}[t] \centering
    \includegraphics[width=\textwidth]{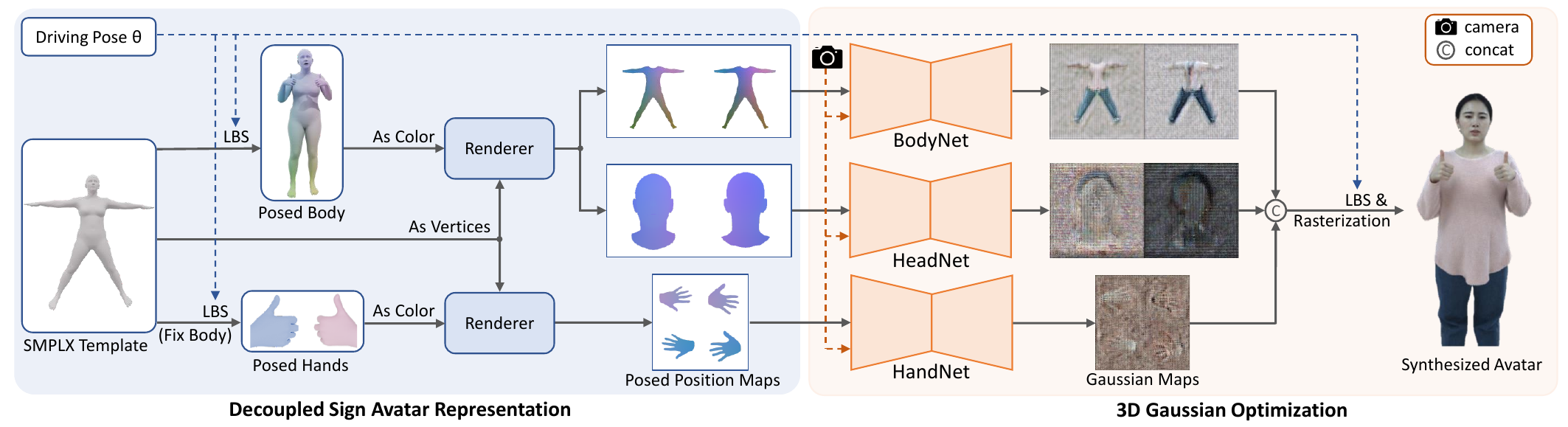}
    \caption{Overview of sign avatar modeling pipeline. The posed vertices are taken as vertex color on the canonical SMPL-X template, and then rendered to generate posed position maps. For the hand part, we employ partial kinematic decoupling by fixing body joints, only preserving hand joint mobility.  We then predict pose-dependent Gaussian maps through three specialized StyleUNet, deform the Gaussians by LBS, and render the synthesized avatar by differentiable rasterization. } \label{fig: avatar_pipeline}
\vspace{-4mm}
\end{figure*}

To improve the quality of sign avatar modeling, we approach the problem from both data and methodological perspectives. From the data perspective, we design a motion-aware sampling strategy that filters out motion-blurred frames and balances the distribution of gestures. On the methodological side, as shown in Figure~\ref{fig: avatar_pipeline}, we introduce a decoupled sign avatar representation that effectively addresses the topological complexity in hand articulation, leading to improved visual quality.

\subsection{\SamplingName}
Motion blur is common in sign language videos, where hand movements are often very fast. Despite increasing the recording frame rate, we observe that motion blur still persists in certain parts of the dataset. Additionally, we identify an imbalance in the types of sign articulations within the dataset. A large portion of the frames contain hand in a stationary, hanging position. This imbalance induces model convergence to suboptimal local minima, substantially compromising the fidelity of hand regions. To address these issues, we design a \SamplingName that filters out motion-blurred frames while also balancing the distribution of sign articulations.
Please refer to the technical appendices for formalized algorithmic workflow.

We first detect motion-blurred frames using three indicators: 1) Hand confidence score from DWPose detection, providing prior information from the pose estimation model. 2) Laplacian gradient of the hand image, which quantifies local texture variations. 3) Hand motion velocity derived from SMPL-X parameter trajectories, serving as a physical constraint. We define thresholds for each indicator and filter out frames that do not meet the criteria set by these indicators.

After filtering out motion-blurred frames, we calculate the distance between each sign gesture based on the SMPL-X parameters. Gestures that are close in distance will be grouped into the same category. We then remove excessive gestures from each class, balancing the number of frames within each sign gesture category.

\subsection{Decoupled Sign Avatar Representation}
\label{sec:avatar_representation}
In sign language, subtle hand gestures and facial expressions are crucial for conveying meaning. 
Previous avatar modeling methods~\cite{li2024animatablegaussians, hu2024gaussianavatar, zhan2025real} typically represent the entire body holistically, which forces the model to learn large-scale body movements together with fine-grained hand and facial motions. 
This joint representation often biases the network toward coarse motion patterns while overlooking the dynamics of the hands and face.

To address this limitation, we propose a decoupled representation that explicitly decomposes the whole-body Gaussians into three anatomically distinct components: body, hands, and head. 
This decomposition enables the model to focus more effectively on each specific region, thereby improving the visual quality. 
Specifically, following ~\cite{li2024animatablegaussians}, we adopt StyleUNet~\cite{wang2023styleavatar} as our backbone to predict pose-dependent Gaussian attributes in canonical space.
To ensure compatibility with 2D networks, the 3D representation of the human avatar needs to be parameterized in 2D space. 
Given a driving pose $\mathbf{\Theta}$, we deform the body and head part through linear blend skinning (LBS). Then we take the posed coordinate as the vertex color on the canonical SMPL-X template, and render it to both front and back views to generate posed position maps $\mathcal{P}$:
\begin{equation}
\label{eq:body_render}
\begin{aligned}
    \mathcal{P}_\text{body}^{\text{f}}, \mathcal{P}_\text{body}^{\text{b}} &= \mathcal{R}(\mathcal{M}_\text{body}, \Psi_\text{LBS}(\mathbf{\Theta}, \mathcal{M}_\text{body})), \\
    \mathcal{P}_\text{head}^{\text{f}}, \mathcal{P}_\text{head}^{\text{b}} &= \mathcal{R}( \mathcal{M}_\text{head}, \Psi_\text{LBS}(\mathbf{\Theta}, \mathcal{M}_\text{head})) ,
\end{aligned}
\end{equation}
where $\mathcal{R}(\mathcal{M}, \mathcal{C})$ denotes the render process of mesh $\mathcal{M}$ with color $\mathcal{C}$, $\Psi_\text{LBS}$ denotes the LBS deformation operator, $\mathcal{M}_\text{body}$ and $\mathcal{M}_\text{head}$ represent the body and head part template mesh.

For hand parameterizations, we further employ partial kinematic decoupling by fixing body joints while preserving hand joint mobility. This isolation ensures that the hand pose maps are only related to hand gestures and are independent of body poses, thus improving the generalization of hand modeling. We render the two hands to both up and down views and concatenate them together:
\begin{equation}
    \mathcal{P}_\text{hand} = \mathcal{R}(\mathcal{M}_\text{hand}, \Psi_\text{LBS}(\mathbf{\Theta}_\text{hand}, \mathcal{M}_\text{hand})),
\end{equation}
where $\mathbf{\Theta}_\text{hand}$ contains hand-specific pose parameters, with the body fixed in its canonical pose.

\subsection{Optimization}
\label{sec:gaussian_optimization}
Our multi-branch architecture employs three specialized StyleUNet modules $\mathcal{F}_\text{body}$, $\mathcal{F}_\text{head}$ and $\mathcal{F}_\text{hand}$ for independent Gaussian attributes prediction:
\begin{equation}
\label{eq:styleunet}
\begin{aligned}
    \mathcal{G}_\text{body}^{\text{f}},\mathcal{G}_\text{body}^{\text{b}} &= \mathcal{F}_\text{body}(\mathcal{P}_\text{body}^{\text{f}}, \mathcal{P}_\text{body}^{\text{b}}, \mathcal{V}), \\
    \mathcal{G}_\text{head}^{\text{f}},\mathcal{G}_\text{head}^{\text{b}} &= \mathcal{F}_\text{head}(\mathcal{P}_\text{head}^{\text{f}}, \mathcal{P}_\text{head}^{\text{b}}, \mathcal{V}), \\
    \mathcal{G}_\text{hand} &= \mathcal{F}_\text{hand}(\mathcal{P}_\text{hand}, \mathcal{V}),
\end{aligned}
\end{equation}
where $\mathcal{G}_\ast$ denotes predicted Gaussian attribute maps (position $\mu$, rotation $r$, scale $s$, opacity $\alpha$, color $c$) in canonical sapce, and $\mathcal{V}$ encodes view-dependent appearance variations.
To ensure that the position attribute of the predicted Gaussian maps closely aligns with the SMPL-X human mesh, we predict position offsets $\Delta\mathcal{P}$ relative to SMPL-X mesh rather than absolute positions.

We then employ LBS to deform the whole body Gaussians from canonical space to observation space. A canonical 3D Gaussian's position $\mu_{c}$ and rotation $r_{c}$ are transformed as follows:
\begin{align}
    \mu_{o} = \mathbf{R} \cdot \mu_{c} + \mathbf{T}, \quad 
    r_{o} = \mathbf{R} \cdot r_{c},
\end{align}
where $\mu_{o}$ and $r_{o}$ are 3D Gaussian's position and rotation in observation space, $\mathbf{R}$ is the rotation matrix and $\mathbf{T}$ is the translation vector, both of which are calculated with the skinning weights of each 3D Gaussian and driving pose.
Finally, we render the posed 3D Gaussians to a desired camera view through splatting-based rasterization~\cite{kerbl20233d}.

\noindent \textbf{Optimization Objectives.}
The composite loss function integrates multiple constraints:
\begin{equation}
\label{eq:loss}
\begin{split}
\mathcal{L} = &\underbrace{\mathcal{L}_{\text{1}} + \lambda_{\text{SSIM}}\mathcal{L}_{\text{SSIM}} + \lambda_{\text{LPIPS}}\mathcal{L}_{\text{LPIPS}}}_{\text{Photometric}} \\
& + \underbrace{\lambda_{\text{hand}}\mathcal{L}_{\text{hand}} + \lambda_{\text{head}}\mathcal{L}_{\text{head}}}_{\text{Anatomical Focus}} + \underbrace{\lambda_{\text{offset}}\|\Delta\mathcal{P}\|_2}_{\text{Regularization}},
\end{split}
\end{equation}
where $\mathcal{L}_{\text{1}}$, $\mathcal{L}_{\text{SSIM}}$, and $\mathcal{L}_{\text{LPIPS}}$ are the L1, SSIM~\cite{wang2004image}, and LPIPS~\cite{zhang2018unreasonable} losses, respectively. $\mathcal{L}_{\text{hand}}$ is the L1 loss on segmented hand regions, encouraging the model to focus on hand details. Similarly, $\mathcal{L}_{\text{head}}$ applies L1 loss on the head region. $\mathcal{L}_{\text{offset}}$ is the L2 norm of position offsets, preventing Gaussians from deviating too far from the template.
\section{Experiments}
\label{sec:experiments}

\subsection{Experimental Setting}
\label{experimental_setting}
\noindent \textbf{Datasets and Metrics.} 
We conduct experiments on our \DatasetName dataset and a set of collected monocular in-the-wild sign language videos. The in-the-wild videos are collected from the Web with the Creative Commons (CC) license. They have four individuals (two males and two females) and are recorded at a resolution of 1920 × 1080. To extract SMPL-X parameters, we apply our hybrid fitting procedure, excluding the 3D triangulation step due to the monocular setting.
To evaluate the results quantitatively, we adopt three commonly-used metrics, including PSNR, SSIM~\cite{wang2004image}, and LPIPS~\cite{zhang2018unreasonable}.

\noindent \textbf{Dataset Split.}
\label{sec:data_split}
For the \DatasetName dataset split, we use the first $\text{90}\%$ of the sampled frames as the training set and the remaining frames as the test set. We use all the 16 views for training.
For in-the-wild sign videos, we uniformly sample the frames with the interval as 1 to split the training and testing frames. The number of training and testing frames are both 150.

\noindent \textbf{Baselines.} 
We compare our method with state-of-the-art human avatar modeling methods,
including  SplattingAvatar~\cite{shao2024splattingavatar}, GaussianAvatar~\cite{hu2024gaussianavatar}, AnimatableGaussians~\cite{li2024animatablegaussians}, EVA~\cite{hu2024expressive} and Mmlphuman~\cite{zhan2025real}, implemented by the official code. Among these methods, EVA focuses on expressive avatar modeling and explicitly accounts for sign language.
Note that recent text-to-video generators for sign language (e.g., SignGen~\cite{qi2024signgen}, SignDiffs~\cite{fang2025signdiff}) follow a 2D image synthesis paradigm with different inputs and do not model a 3D drivable avatar, thus they are not directly comparable.
Since some of the baselines~\cite{shao2024splattingavatar, hu2024gaussianavatar} are animated with SMPL~\cite{SMPL:2015} which has no control over the hands and facial expressions.
We replace the driven model with SMPL-X~\cite{SMPL-X:2019} and increase the pose-conditioning dimension accordingly, while keeping network, losses, training schedule, and hyperparameters unchanged. For fairness, all methods use the same train/test split, camera views, images, fitted SMPL-X supervision, and masks.

\noindent \textbf{Implementation Details.} 
Our framework is implemented with PyTorch and all experiments are performed on a single NVIDIA 3090 GPU. We set the optimization hyperparameters $\lambda_{\text{SSIM}}=0.1$, $\lambda_{\text{LPIPS}}=0.1$, $\lambda_{\text{hand}}=3$, $\lambda_{\text{head}}=3$ and $\lambda_{\text{offset}}=0.005$. The resolution of the posed position maps for body, hand and head part are $1024\times 512$, $256\times 256$ and $256\times 128$, respectively. Please refer to the supplementary material for more details.

\subsection{Dataset Analysis}
\label{ablation}
Unlike previous multi-view human-centric~\cite{cheng2023dna, icsik2023humanrf, peng2021neural, xiong2024mvhumannet} or sign language datasets~\cite{yu2024signavatars, PHOENIX14t, Duarte_CVPR2021}, \DatasetName captures both multi-view imagery and diverse sign gestures, which are crucial for learning expressive and generalizable sign avatars.
As shown in Table~\ref{tab:ablation_dataset_full} and Figure~\ref{fig:abla_data_view_num}, increasing the number of views enhances visual quality, especially for fine-grained hand details. This demonstrates that multi-view supervision is essential for resolving self-occlusions and depth ambiguities that are inherent in complex hand gestures.
We further analyze the effect of different signing patterns in \DatasetName. As shown in Table~\ref{tab:ablation_dataset_full2}, training on only the basic hand shapes subset limits generalization due to the lack of diverse movement patterns. Conversely, training solely on sign sentences makes the learning process more challenging, leading to degraded quality. Combining both subsets resulting in a more robust and realistic sign avatar. These experiments highlight the necessity of \DatasetName multi-view and diverse sign articulations, distinguishing it from existing sign language datasets.

\begin{table}[t] \centering
    \scriptsize
    \setlength{\tabcolsep}{3pt}
    \caption{Ablation study of hybrid SMPL-X fitting strategy. “↓” indicates that lower values are better, while “↑” means the opposite.}
    \resizebox{\textwidth}{!}{%
        \begin{tabular}{l ccc ccc ccc}
\toprule
\multirow{2}{*}{\textbf{Method}}  & \multicolumn{3}{c}{\textbf{Full}\quad\quad} & \multicolumn{3}{c}{\textbf{Hand}\quad\quad} & \multicolumn{3}{c}{\textbf{Face}\quad\quad} \\ 
\cmidrule(lr){2-4} \cmidrule(lr){5-7} \cmidrule(lr){8-10}
& {\bf PSNR}$\uparrow$ & {\bf SSIM}$\uparrow$ & {\bf LPIPS}$\downarrow$ & {\bf PSNR}$\uparrow$ & {\bf SSIM}$\uparrow$ & {\bf LPIPS}$\downarrow$ & {\bf PSNR}$\uparrow$ & {\bf SSIM}$\uparrow$ & {\bf LPIPS}$\downarrow$ \\ \midrule
{w/o Hybrid Fitting} & 23.58 & 0.9637 & 0.0618 & 16.82 & 0.7371 & 0.2987 & 16.62 & 0.7923 & 0.2435 \\
{w Hybrid Fitting} & {\bf 26.90} & {\bf 0.9722} & {\bf 0.0370} & {\bf 18.85} & {\bf 0.7704} & {\bf0.2301} & {\bf 20.51} & {\bf 0.8499} & {\bf 0.1665} \\
\bottomrule
\end{tabular}
    }
    \label{tab:ablation_smplx}
\end{table}

\begin{figure}[t]
\centering
\begin{minipage}[t]{0.49\linewidth}
    \vspace{2mm}
    \centering
    \includegraphics[width=\linewidth]{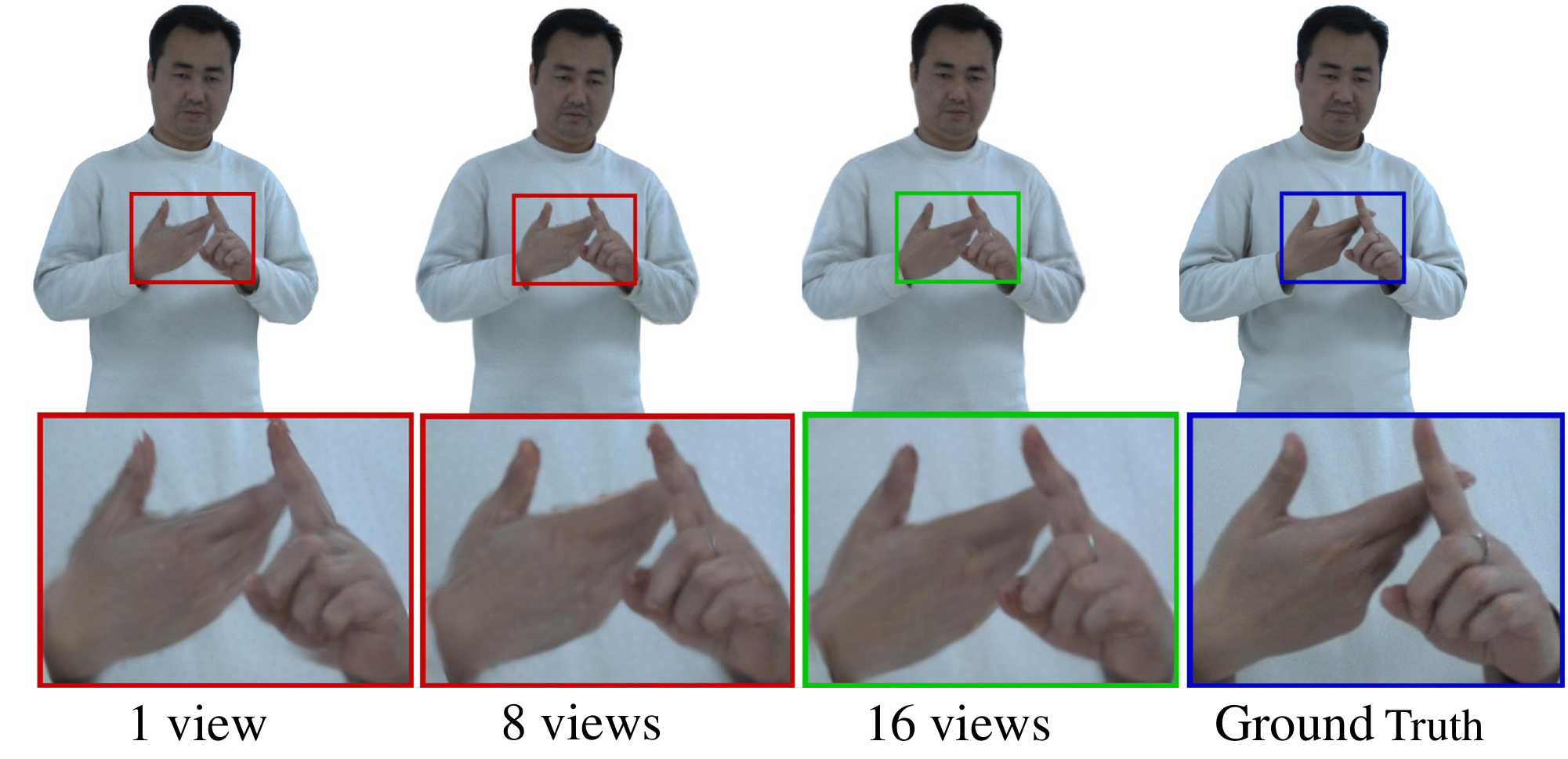}
    \caption{Visualization of animated avatar trained with different numbers of views.}
    \label{fig:abla_data_view_num}
\end{minipage}\hfill
\begin{minipage}[t]{0.49\linewidth}
    \vspace{0pt}
    \centering
    \footnotesize
    \setlength{\tabcolsep}{10pt}

    \captionof{table}{Comparison with other SMPL-X estimation methods on the SGNify~\cite{forte2023reconstructing} mocap dataset. “↓” indicates that lower values are better.}
    \label{tab:img2smpl_compare}
    \vspace{1mm}
    \resizebox{\linewidth}{!}{%
        \begin{tabular}{lccc}
\toprule
\textbf{Method} & \textbf{Body}$\downarrow$ & \textbf{Left Hand}$\downarrow$ & \textbf{Right Hand}$\downarrow$ \\
\midrule
FrankMoCap~\cite{rong2021frankmocap} & 78.07 & 20.47 & 19.62 \\
PIXIE~\cite{feng2021collaborative}           & 60.11 & 25.02 & 22.42 \\
PyMAF-X~\cite{zhang2023pymaf}       & 68.61 & 21.46 & 19.19 \\
SMPLify-X~\cite{SMPL-X:2019}   & 56.07 & 22.23 & 18.83 \\
SGNify~\cite{forte2023reconstructing}         & 55.63 & 19.22 & 17.50 \\
NSA~\cite{baltatzis2024neural} & 46.42 & 16.17 & 15.23 \\
\textbf{Ours}            & \textbf{40.38} & \textbf{14.11} & \textbf{13.79} \\
\bottomrule
\end{tabular}

    }
\end{minipage}
\vspace{-3mm}
\end{figure}

\begin{table}[t] \centering
    \scriptsize
    \setlength{\tabcolsep}{3pt}
    \caption{Impact of the number of views on \DatasetName dataset. “↓” indicates that lower values are better, while “↑” means the opposite.}
    \resizebox{\textwidth}{!}{%
        \begin{tabular}{l ccc ccc ccc}
\toprule
\multirow{2}{*}{\textbf{Method}}  & \multicolumn{3}{c}{\textbf{Full}\quad\quad} & \multicolumn{3}{c}{\textbf{Hand}\quad\quad} & \multicolumn{3}{c}{\textbf{Face}\quad\quad} \\ 
\cmidrule(lr){2-4} \cmidrule(lr){5-7} \cmidrule(lr){8-10}
& {\bf PSNR}$\uparrow$ & {\bf SSIM}$\uparrow$ & {\bf LPIPS}$\downarrow$ & {\bf PSNR}$\uparrow$ & {\bf SSIM}$\uparrow$ & {\bf LPIPS}$\downarrow$ & {\bf PSNR}$\uparrow$ & {\bf SSIM}$\uparrow$ & {\bf LPIPS}$\downarrow$ \\ \midrule
{1} & 26.17 & 0.9711 & 0.0425 & 18.49 & 0.7599 & 0.2581 & 18.84 & 0.8223 & 0.1927 \\
{8} & 26.42 & 0.9711 & 0.0417 & 18.60 & 0.7649 & 0.2524 & 19.43 & 0.8286 & 0.1759 \\
Full (16) & {\bf 26.90} & {\bf 0.9722} & {\bf 0.0370} & {\bf 18.85} & {\bf 0.7704} & {\bf0.2301} & {\bf 20.51} & {\bf 0.8499} & {\bf 0.1665} \\
\bottomrule
\end{tabular}
    }
    \label{tab:ablation_dataset_full}
\end{table}

\begin{table}[t] \centering
    \scriptsize
    \setlength{\tabcolsep}{3pt}
    \caption{Impact of incorporated signing patterns on \DatasetName dataset. “↓” indicates that lower values are better, while “↑” means the opposite.}
    \resizebox{\textwidth}{!}{%
        \begin{tabular}{l ccc ccc ccc}
\toprule
\multirow{2}{*}{\textbf{Method}}  & \multicolumn{3}{c}{\textbf{Full}\quad\quad} & \multicolumn{3}{c}{\textbf{Hand}\quad\quad} & \multicolumn{3}{c}{\textbf{Face}\quad\quad} \\ 
\cmidrule(lr){2-4} \cmidrule(lr){5-7} \cmidrule(lr){8-10}
& {\bf PSNR}$\uparrow$ & {\bf SSIM}$\uparrow$ & {\bf LPIPS}$\downarrow$ & {\bf PSNR}$\uparrow$ & {\bf SSIM}$\uparrow$ & {\bf LPIPS}$\downarrow$ & {\bf PSNR}$\uparrow$ & {\bf SSIM}$\uparrow$ & {\bf LPIPS}$\downarrow$ \\ \midrule
{Only Basic Hand Shapes} & 24.08 & 0.9660 & 0.0524 & 17.47 & 0.7425 & 0.2892 & 17.55 & 0.8034 & 0.2287 \\
{Only Sign Sentences} & 24.36 & 0.9653 & 0.0501 & 17.62 & 0.7518 & 0.2768 & 17.76 & 0.8122 & 0.2139 \\
% \midrule
Ours & {\bf 26.90} & {\bf 0.9722} & {\bf 0.0370} & {\bf 18.85} & {\bf 0.7704} & {\bf0.2301} & {\bf 20.51} & {\bf 0.8499} & {\bf 0.1665} \\
\bottomrule
\end{tabular}

    }
    \label{tab:ablation_dataset_full2}
\end{table}

\subsection{Ablation Study}

\noindent \textbf{Effectiveness of hybrid SMPL-X fitting.}
We evaluate our hybrid SMPL-X fitting strategy from two perspectives: accuracy and its effect on modeling.
To assess accuracy, we compare our pipeline against prior methods, including SGNify~\cite{forte2023reconstructing} and Neural Sign Actors (NSA)~\cite{baltatzis2024neural}, both designed for SMPL-X estimation in sign language contexts. The evaluation is conducted on the SGNify mocap dataset following its official protocol, measuring the mean per-vertex error (mm) on the upper body, left hand, and right hand. As shown in Table~\ref{tab:img2smpl_compare}, our method achieves the lowest error, particularly for the hands, demonstrating its superior ability to capture precise hand motions.
To evaluate its effect on avatar modeling, we compare models trained using SMPL-X parameters fitted from DWPose 2D keypoints versus our hybrid pipeline. As shown in Table~\ref{tab:ablation_smplx}, directly fitting from DWPose 2D keypoints fails to capture complex hand motions, whereas our hybrid fitting provides accurate hand poses, resulting in fine-grained hand details.

\begin{figure}[H]
\centering
\begin{minipage}[t]{0.49\linewidth}
    \vspace{0pt}
    \centering
    \includegraphics[width=\linewidth]{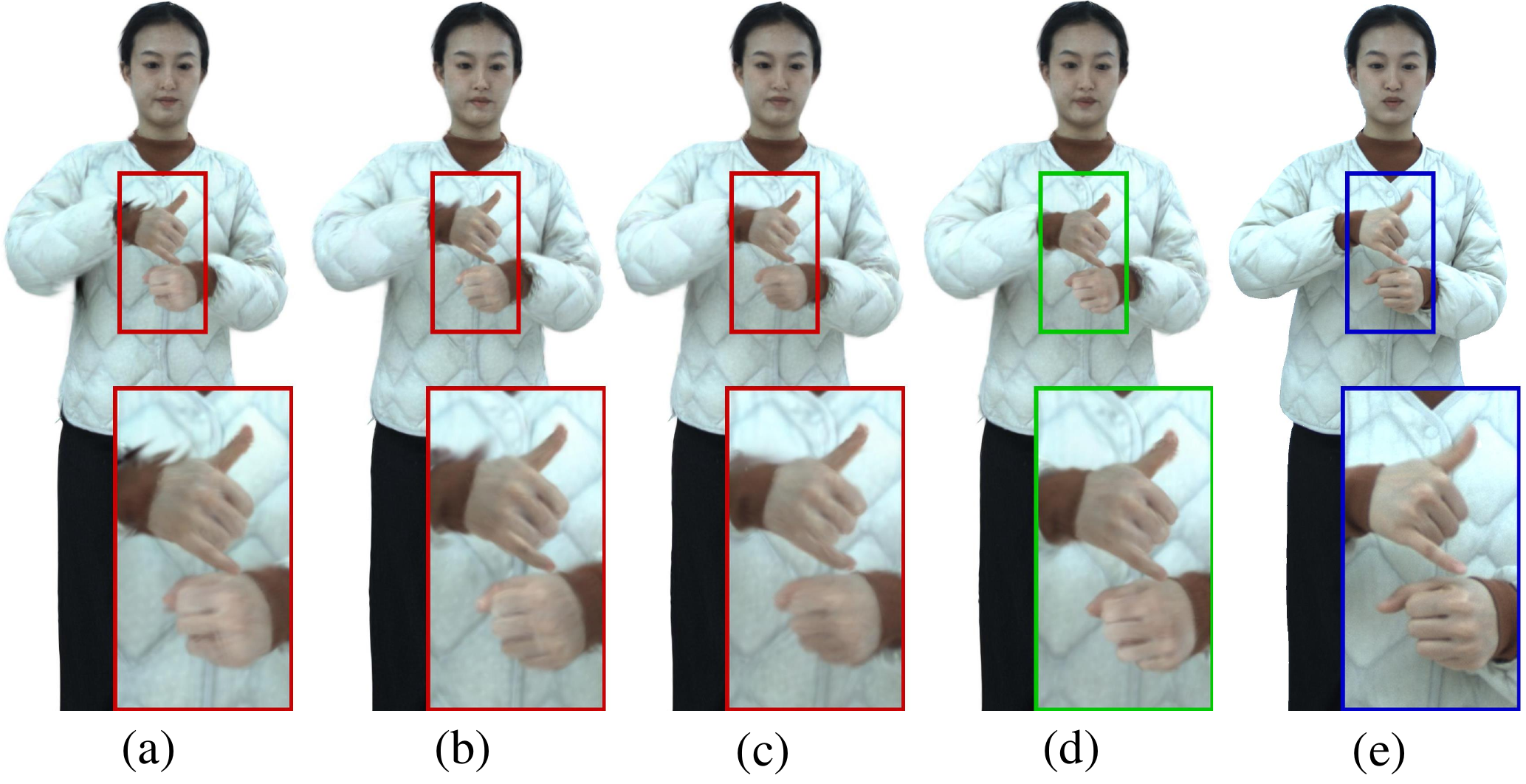}
    \caption{Ablation study of sampling strategy. (a) Sequential sampling; (b) Isometric sampling; (c) Random sampling; (d) Our motion-aware sampling; (e) Ground truth image.}
    \label{fig:abla_sample}
\end{minipage}\hfill
\begin{minipage}[t]{0.49\linewidth}
    \vspace{0pt}
    \centering
    \includegraphics[width=\linewidth]{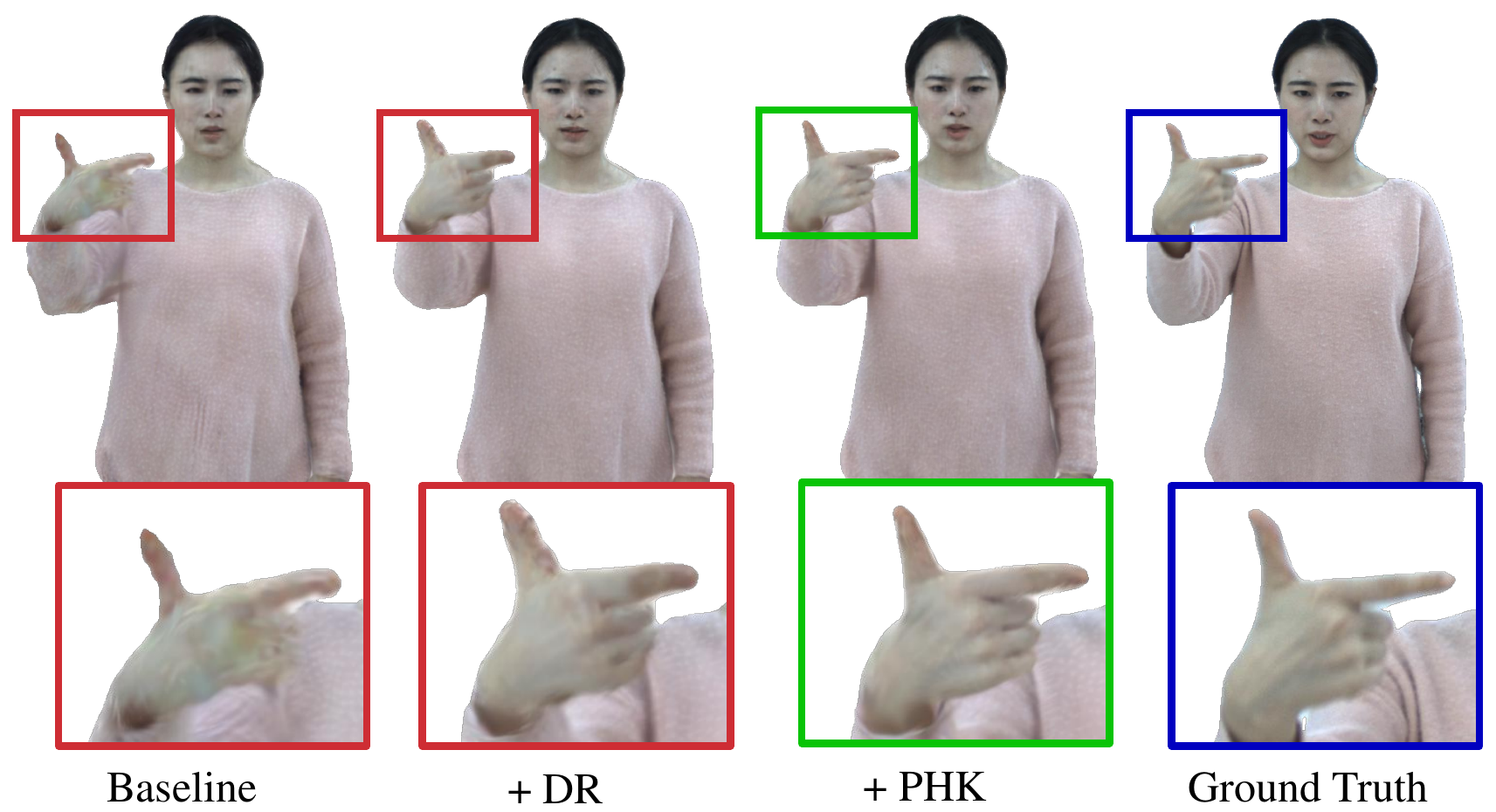}
    \caption{Visualization of animated avatar trained with different representation. “DR” indicates the decoupled representation, while “PHK” means Partial Hand Kinematic.}
    \label{fig:ablation_decouple}
\end{minipage}
\vspace{-3mm}
\end{figure}

\noindent \textbf{Effect of the Data Sampling Strategy.} We evaluate four sampling strategies: sequential, isometric, random, and our proposed \SamplingName to assess how frame selection affects modeling. As shown in Table~\ref{tab:ablation_sample_strategy} and Figure~\ref{fig:abla_sample}, filtering out invalid frames using our sampling strategy leads to improved visual quality. In contrast, other strategies often include motion-blurred frames, and overfitting to these low-quality frames degrades the accuracy of fine-grained hand details.

\noindent \textbf{Effects of Decoupled Design.} 
Starting from a single Gaussian map baseline~\cite{li2024animatablegaussians}, we evaluate decoupling the representation into body, head, and hands, and further applying partial kinematic decoupling for hands. As shown in Table~\ref{tab:ablation_decouple} and Figure~\ref{fig:ablation_decouple}, decoupling alone gives minor gains, whereas incorporating partial kinematic decoupling significantly enhances hand quality, underscoring the importance of fully decoupling hand representations.

\begin{table}[t] \centering
    \scriptsize
    \setlength{\tabcolsep}{3pt}
    \caption{Ablation study of data sampling strategy. “↓” indicates that lower values are better, while “↑” means the opposite.}
    \resizebox{\textwidth}{!}{%
        \begin{tabular}{l ccc ccc ccc}
\toprule
\multirow{2}{*}{\textbf{Strategy}}  & \multicolumn{3}{c}{\textbf{Full}\quad\quad} & \multicolumn{3}{c}{\textbf{Hand}\quad\quad} & \multicolumn{3}{c}{\textbf{Face}\quad\quad} \\ 
\cmidrule(lr){2-4} \cmidrule(lr){5-7} \cmidrule(lr){8-10}
& {\bf PSNR}$\uparrow$ & {\bf SSIM}$\uparrow$ & {\bf LPIPS}$\downarrow$ & {\bf PSNR}$\uparrow$ & {\bf SSIM}$\uparrow$ & {\bf LPIPS}$\downarrow$ & {\bf PSNR}$\uparrow$ & {\bf SSIM}$\uparrow$ & {\bf LPIPS}$\downarrow$ \\ \midrule
{Sequential} & 26.55 & 0.9659 & 0.0372 & 16.40 & 0.6734 & 0.2656 & 18.66 & 0.7733 & 0.1875 \\
{Isometric} & 26.47 & 0.9659 & 0.0371 & 16.29 & 0.6697 & 0.2686 & 18.57 & 0.7726 & 0.1906 \\
{Random} & 26.90 & 0.9663 & 0.0353 & 17.40 & 0.6939 & 0.2490 & 18.94 & 0.7838 & 0.1784 \\
{Motion-aware} & {\bf 27.67} & {\bf 0.9684} & {\bf 0.0340} & {\bf 18.03} & {\bf 0.7161} & {\bf0.2387} & {\bf 19.79} & {\bf 0.8029} & {\bf 0.1672} \\
\bottomrule
\end{tabular}
    }
    \label{tab:ablation_sample_strategy}
\vspace{-2mm}
\end{table}

\begin{table}[t] \centering
    \scriptsize
    \setlength{\tabcolsep}{3pt}
    \caption{Ablation study of decoupled sign avatar representation. “DR” indicates the decoupled representation, while “PHK” means Partial Hand Kinematic. “↓” indicates that lower values are better, while “↑” means the opposite.}
    \resizebox{\textwidth}{!}{%
        \begin{tabular}{l ccc ccc ccc}
\toprule
\multirow{2}{*}{\textbf{Method}}  & \multicolumn{3}{c}{\textbf{Full}\quad\quad} & \multicolumn{3}{c}{\textbf{Hand}\quad\quad} & \multicolumn{3}{c}{\textbf{Face}\quad\quad} \\ 
\cmidrule(lr){2-4} \cmidrule(lr){5-7} \cmidrule(lr){8-10}
& {\bf PSNR}$\uparrow$ & {\bf SSIM}$\uparrow$ & {\bf LPIPS}$\downarrow$ & {\bf PSNR}$\uparrow$ & {\bf SSIM}$\uparrow$ & {\bf LPIPS}$\downarrow$ & {\bf PSNR}$\uparrow$ & {\bf SSIM}$\uparrow$ & {\bf LPIPS}$\downarrow$ \\ \midrule
{Baseline} & 25.24 & 0.9570 & 0.0490 & 16.65 & 0.6912 & 0.3759 & 18.47 & 0.7948 & 0.2167 \\
{+ DR} & 26.28 & 0.9607 & 0.0430 & 17.53 & 0.7174 & 0.3188 & 20.53 & 0.8099 & 0.1731 \\
{+ PHK} & {\bf 27.93} & {\bf 0.9665} & {\bf 0.0388} & {\bf 18.74} & {\bf 0.7322} & {\bf 0.2666} & {\bf 21.50} & {\bf 0.8253} & {\bf 0.1595} \\
\bottomrule
\end{tabular}
    }
    \label{tab:ablation_decouple}
\end{table}

\begin{figure}[t]
\centering
\begin{minipage}[t]{0.49\linewidth}
    \vspace{0pt}
    \centering
    \footnotesize
    \captionof{table}{User study results comparing SMPL-X mesh rendering and our photorealistic sign avatar.}
    \label{tab:user_study_smplx}
    \vspace{1mm}
    \resizebox{\linewidth}{!}{%
        \begin{tabular}{lcc}
\toprule
\textbf{Representation} & SMPL-X Mesh & \textbf{Photorealistic Sign Avatar} \\
\midrule
Comprehensibility & 17.5\% & \textbf{82.5\%} \\
Hand/Facial Clarity & 33.5\% & \textbf{66.5\%} \\
Visual Realism & 5.0\% & \textbf{95.0\%} \\
Aesthetic Preference & 16.0\% & \textbf{84.0\%} \\
\bottomrule
\end{tabular}
    }
\end{minipage}\hfill
\begin{minipage}[t]{0.49\linewidth}
    \vspace{0pt}
    \centering
    \footnotesize
    \captionof{table}{User study results comparing our method with other human avatar modeling methods.}
    \label{tab:user_study}
    \vspace{1mm}
    \resizebox{\linewidth}{!}{%
        \begin{tabular}{lcccc}
\toprule
\textbf{Method} & AnimatableGS~\cite{li2024animatablegaussians} & EVA~\cite{hu2024expressive} & Mmlphuman~\cite{zhan2025real} & \textbf{Ours} \\
\midrule
Comprehensibility & 8.8\% & 13.6\% & 20.8\% & \textbf{56.8\%} \\
Hand/Facial Clarity & 7.8\% & 11.2\% & 18.6\% & \textbf{62.4\%} \\
Temporal Consistency & 8.0\% & 13.8\% & 25.2\% & \textbf{53.0\%} \\
Aesthetic Preference & 11.2\% & 15.6\% & 22.0\% & \textbf{51.2\%} \\
\bottomrule
\end{tabular}
    }
\end{minipage}
\vspace{-3mm}
\end{figure}

\begin{table*}[t] \centering
    \scriptsize
    \setlength{\tabcolsep}{3pt}
    \renewcommand{\arraystretch}{0.98}
    \caption{Comparison to state-of-the-art human avatar modeling methods. “↓” indicates that lower values are better, while “↑” means the opposite.}
    \resizebox{\linewidth}{!}{%
        \begin{tabular}{l ccc ccc ccc}
\toprule
\multirow{2}{*}{\textbf{Method}}  & \multicolumn{3}{c}{\textbf{Full}\quad\quad} & \multicolumn{3}{c}{\textbf{Hand}\quad\quad} & \multicolumn{3}{c}{\textbf{Face}\quad\quad} \\ 
\cmidrule(lr){2-4} \cmidrule(lr){5-7} \cmidrule(lr){8-10}
& {\bf PSNR}$\uparrow$ & {\bf SSIM}$\uparrow$ & {\bf LPIPS}$\downarrow$ & {\bf PSNR}$\uparrow$ & {\bf SSIM}$\uparrow$ & {\bf LPIPS}$\downarrow$ & {\bf PSNR}$\uparrow$ & {\bf SSIM}$\uparrow$ & {\bf LPIPS}$\downarrow$ \\ \midrule
\multicolumn{10}{l}{\emph{Multi-view setting: MVSign dataset}} \\
{SplattingAvatar~\cite{shao2024splattingavatar}} & 23.71 & 0.9625 & 0.0494 & 15.64 & 0.6762 & 0.3884 & 16.90 & 0.7633 & 0.2423 \\
{GaussianAvatar~\cite{hu2024gaussianavatar}} & 24.23 & 0.9633 & 0.0428 & 16.30 & 0.6833 & 0.3082 & 17.78 & 0.7737 & 0.2086 \\
{AnimatableGS~\cite{li2024animatablegaussians}} & 25.09 & 0.9647 & 0.0465 & 16.95 & 0.7002 & 0.3135 & 18.63 & 0.7842 & 0.2230 \\
{EVA~\cite{hu2024expressive}} & 25.56 & 0.9667 & 0.0436 & 17.43 & 0.7154 & 0.2869 & 19.21 & 0.8000 & 0.1964 \\
{Mmlphuman~\cite{zhan2025real}} & 25.91 & 0.9652 & 0.0460 & 17.41 & 0.7084 & 0.2675 & 19.38 & 0.8025 & 0.1869 \\
{Ours} & {\bf 27.03} & {\bf 0.9689} & {\bf 0.0393} & {\bf 18.55} & {\bf 0.7325} & {\bf 0.2568} & {\bf 20.60} & {\bf 0.8189} & {\bf 0.1757} \\
\midrule
\multicolumn{10}{l}{\emph{Monocular setting: Real-world Web video}} \\
{SplattingAvatar~\cite{shao2024splattingavatar}} & 18.71 & 0.9198 & 0.1223 & 16.37 & 0.6572 & 0.3114 & 16.97 & 0.7243 & 0.2751 \\
{GaussianAvatar~\cite{hu2024gaussianavatar}} & 18.90 & 0.9261 & 0.0918 & 16.30 & 0.6668 & 0.2797 & 17.29 & 0.7178 & 0.2157 \\
{AnimatableGS~\cite{li2024animatablegaussians}} & 19.34 & 0.9252 & 0.0954 & 17.42 & 0.6951 & 0.2799 & 17.65 & 0.7277 & 0.2561  \\
{EVA~\cite{hu2024expressive}} & 19.22 & 0.9250 & 0.0966 & 17.49 & 0.6958 & 0.2588 & 17.75 & 0.7381 & 0.2132 \\
{Mmlphuman~\cite{zhan2025real}} & 19.61 & 0.9239 & 0.1023 & 16.90 & 0.6645 & 0.2607 & 17.88 & 0.7284 & 0.2374 \\
{Ours} & {\bf 20.39} & {\bf 0.9334} & {\bf 0.0773} & {\bf 18.70} & {\bf 0.7189} & {\bf 0.2289} & {\bf 18.84} & {\bf 0.7502} & {\bf 0.1864} \\
\bottomrule
\end{tabular}
    }
    \label{tab:main_table}
\vspace{-2mm}
\end{table*}

\begin{figure*}[t]
\centering
\includegraphics[width=\textwidth,]{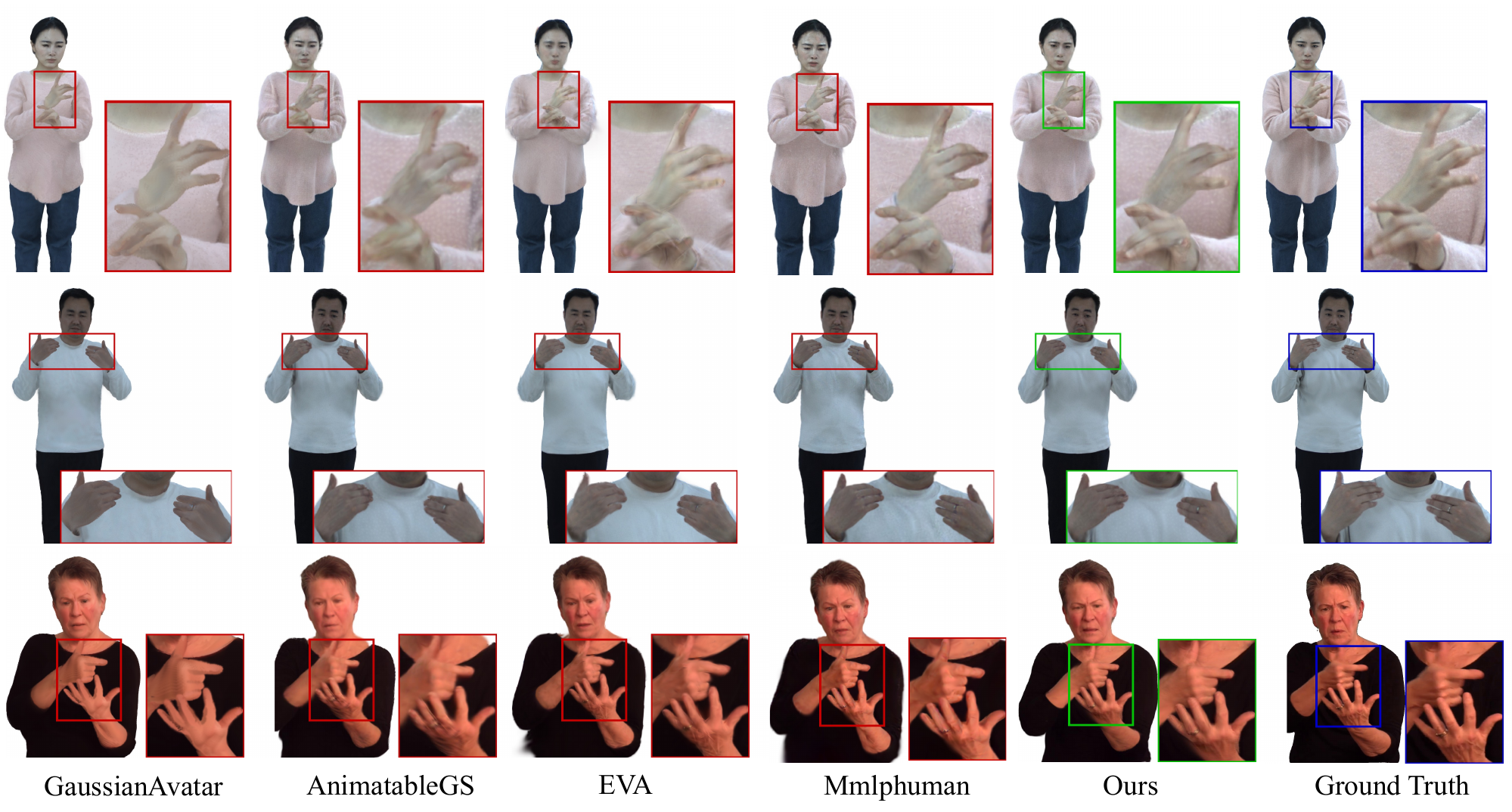}
\caption{Qualitative comparison of novel pose synthesis with GaussianAvatar~\cite{hu2024gaussianavatar}, AnimatableGS~\cite{li2024animatablegaussians}, EVA~\cite{hu2024expressive} and Mmlphuman~\cite{zhan2025real}. The first two rows show results on the \DatasetName dataset, while the last row presents results on real-world web videos.}
\label{fig:compare_testset}
\vspace{-4mm}
\end{figure*}

\subsection{Comparison with State-of-the-art Methods}
\label{comparison}

\noindent \textbf{Quantitative and Qualitative Comparison.}
As shown in Table~\ref{tab:main_table}, our method outperforms the baselines across all evaluation metrics.
Specifically, on the \DatasetName dataset, it achieves an improvement of $\text{8.1}\%$, $\text{4.0}\%$ and $\text{5.9}\%$ relative LPIPS gain on the full, hand and face regions, respectively.
On the in-the-wild sign videos, the performance gains are more pronounced, with $\text{15.8}\%$, $\text{11.6}\%$ and $\text{12.6}\%$ relative LPIPS improvements on the full, hand and face regions.
These results demonstrate that our method not only performs strongly on the \DatasetName dataset but also generalizes effectively to complex real-world sign language videos.
The qualitative comparisons in Figure~\ref{fig:compare_testset} show that baselines struggle with complex hand gestures, producing blurred or coarse hands. In contrast, our decoupled avatar representation captures intricate hand motions, yielding more realistic results.

\noindent \textbf{User Study.}
In addition to signal-based metrics, we conducted a user study with native Deaf people to evaluate perceptual quality. 
We distributed 20 survey forms via Deaf community channels and collected 20 valid responses.
Participants evaluated 25 sets of videos generated by three representative baseline methods~\cite{li2024animatablegaussians, hu2024expressive, zhan2025real} and ours.
For each set, participants were asked to select the single video that best satisfied the corresponding criterion:
\begin{itemize}
    \item \textbf{Q1:} Which video best allows you to understand the intended sign-language content? 
    (\emph{Comprehensibility})
    
    \item \textbf{Q2:} Which video best presents clear hand gestures and facial expressions?
    (\emph{Hand/Facial Clarity})

    \item \textbf{Q3:} Which video shows the most temporally stable visual appearance?
    (\emph{Temporal Consistency}) 

    \item \textbf{Q4:} Which video is most visually appealing and acceptable as a sign-language avatar?
    (\emph{Aesthetic Preference})
\end{itemize}
As shown in Table~\ref{tab:user_study}, our method achieves the highest ratings across all four aspects, indicating superior perceptual quality from the participants' perspective.

We further conduct a user study within the Deaf community to evaluate the perceptual preference between our photorealistic sign avatar and the SMPL-X mesh. 
We distributed 20 survey forms via Deaf community channels and collected 20 valid responses. Participants evaluated 10 video pairs, which compare SMPL-X mesh rendering against our photorealistic avatar under the same motion, and rated them from four aspects: comprehensibility, hand/facial clarity, visual realism, and aesthetic preference.
As reported in Table~\ref{tab:user_study_smplx}, participants consistently preferred our photorealistic avatar across these aspects (with 84\% favoring ours in aesthetic preference), showing a clear advantage in perceptual quality and user acceptance.
\section{Limitations}
\label{sec:limitations}
Since we employ three specialized StyleUNet modules for Gaussian attribute prediction, inference speed is largely dominated by their forward passes. This computational overhead limits real-time applicability. A promising direction is replacing StyleUNet with lightweight architectures to improve efficiency while maintaining acceptable visual fidelity.

\section{Conclusion}
\label{sec:conclusions}
In this work, we focus on photorealistic and drivable sign avatar modeling. To this end, we introduce \DatasetName, the first multi-view Chinese sign language dataset co-designed with Deaf experts, featuring diverse gestures and rich annotations.
Our dataset includes comprehensive annotations along with the raw data, including image matting, body part segmentation, 3D keypoints, and SMPL-X parameters.
For precise SMPL-X annotation, we propose a hybrid SMPL-X fitting pipeline that integrates outputs from multiple models, producing accurate body, hand, and facial parameters and can also be applied to the monocular setting.
Building on \DatasetName, we introduce a decoupled sign avatar representation that addresses the topological complexity in hand articulation, while a motion-aware data sampling strategy is applied to filter out motion-blurred frames while balancing the distribution of sign articulations.
Extensive experiments on \DatasetName dataset and in-the-wild sign videos demonstrate that our method achieves strong performance both quantitatively and qualitatively.

\section{Acknowledgments}
\label{sec:acknowledgement}
This work was supported by the Youth Innovation Promotion Association CAS. It was also supported by the GPU cluster built by MCC Lab of Information Science and Technology Institution, USTC, and the Supercomputing Center of USTC.

\bibliographystyle{splncs04}
\bibliography{main}
\end{document}